\documentclass{article}

\usepackage{PRIMEarxiv}

\usepackage[utf8]{inputenc} 
\usepackage[T1]{fontenc}    
\usepackage{hyperref}       
\usepackage{url}            
\usepackage{booktabs}       
\usepackage{caption}
\usepackage{xcolor}
\usepackage{multirow}
\usepackage{amsfonts}       
\usepackage{subcaption}
\usepackage{amsmath}
\usepackage[numbers]{natbib}
\usepackage{nicefrac}       
\usepackage{microtype}      
\usepackage{lipsum}
\usepackage{fancyhdr}       
\usepackage{graphicx}       
\graphicspath{{images/}}    

\title{TrafFormer: A Transformer Model for Predicting Long-Term Traffic
}

\author{
  David Alexander Tedjopurnomo \\
  RMIT University \\
  Melbourne, Australia \\
  \texttt{dtedjopurnom@gmail.com} \\
  \And
  Farhana M. Choudhury \\
  The University of Melbourne \\
  Parkville, Australia\\
  \texttt{farhana.choudhury@unimelb.edu.au} \\
  \And
  A. K. Qin \\
  Swinburne University of Technology \\
  Hawthorn, Australia\\
  \texttt{kqin@swin.edu.au} \\
}

\begin{document}
\newcommand{\bao}[1]{}
\newcommand{\david}[1]{}
\newcommand{\kai}[1]{}
\newcommand{\farhana}[1]{}

\newcommand{\cmark}{\ding{51}}
\newcommand{\xmark}{\ding{55}}
\newcommand{\adv}{A(s,a;\theta_\pi,\theta_Q)}

\newcommand{\numpapers}{48 }
\newcommand{\trafficy}{\scalebox{2.2}{$\scriptscriptstyle y$}}

\newcommand{\db}{$\mathcal{Y}$ }
\newcommand{\dbmath}{\mathcal{Y} }
\newcommand{\Ell}[1]{$\mathcal{L}_{#1}$}
\newcommand{\norm}[1]{\left\lVert #1 \right\rVert_2}
\newcommand{\query}{$\mathcal{X}$ }
\newcommand{\sspan}{$\delta_p$ }
\newcommand{\stride}{$\delta_t$ }

\newcommand{\cee}{\mathbb{C}}%
\newcommand{\iii}{\mathbb{I}}%
\newcommand{\meaneff}{\overline{\cee_I}}
\newcommand{\meaneffplus}{\overline{\cee_{I^+}}}
\newcommand{\meaneffminus}{\overline{\cee_{I^-}}}
\maketitle

\begin{abstract}\label{sec:0_abstract}
Traffic prediction is a flourishing research field due to its importance in human mobility in the urban space. Despite this, existing studies only focus on short-term prediction of up to few hours in advance, with most being up to one hour only. Long-term traffic prediction can enable more comprehensive, informed, and proactive measures against traffic congestion and is therefore an important task to explore. In this paper, we explore the task of long-term traffic prediction; where we predict traffic up to 24 hours in advance. We note the weaknesses of existing models--which are based on recurrent structures--for long-term traffic prediction and propose a modified Transformer model "TrafFormer". Experiments comparing our model with existing hybrid neural network models show the superiority of our model. 
\end{abstract}
\section{Introduction}\label{sec:1_introduction}

Traffic congestion is a major problem faced by metropolitan cities. In 2015, it is estimated that the avoidable cost of traffic congestion for Australian capital cities is approximately \$16.5 billion, up from the 2010 estimate of \$12.8 billion. Furthermore, this value is estimated to increase to about \$30 billion by 2030 \cite{cosgove2015traffic}. Most congestion mitigation measures are costly, difficult to implement, or both. For instance, Singapore implemented regulations on the number of vehicles on roads \cite{BBC2017}, which is infeasible for countries with poor public transportation systems. Constructing new roads to ease congestion is also difficult due to the extremely high cost. As an example, the estimated per mile cost of a standard one-lane road in New Jersey, USA is \$220,490 \cite{Carnegie2016}. 

With the advancements and widespread adoption of traffic sensors, access to large traffic databases is now available. This has led to the development of traffic prediction as a research field starting from the late 70s. Educated traffic decision made through accurate prediction is a far cheaper and easier-to-implement alternative for reducing road congestion. Future traffic prediction involves creating a prediction model from historical traffic data to predict the short-term future traffic ranging from 5 to 60 minutes into the future. These traffic data are commonly collected by traffic induction loop detectors, which detect the presence of passing vehicles. The most common values to be recorded by these detectors are traffic speed and traffic flow, which are defined as the average car speed and the total number of vehicles passing through the loop detector within a period of time, respectively. 

Traffic prediction is different from conventional time-series analysis in that traffic prediction is subject to spatial factors as well as many other external factors. For instance, the prediction of traffic at one site depends on the traffic at other sites and all of the sites are affected by external factors such as weather and holidays. Amongst all the available traffic prediction methods, deep neural network is the most prominent. This is due to its sheer predictive power that can model the complex and nonlinear traffic patterns \cite{YuHaiyang2017}\cite{Huang2014}\cite{Wu2016}\cite{Cheng2017}\cite{Dai2017}. The three most common deep neural network models used for traffic prediction are Convolutional Neural Networks, Recurrent Neural Networks, and Feedforward Neural Networks. 

Traffic prediction is a well-studied research field, but current research only focuses on short-term prediction, usually for up to 60 minutes in advance. Longer-term prediction for up to 24 hours ahead has not been considered despite its usefulness, as longer-term prediction allows public transport planners to take better proactive measures against traffic congestion. In addition, the existing body of literature uses a hybrid approach that relies on recurrent structures to model the traffic data. This can work on short-term prediction, but on long prediction horizons, the instability of the error propagation may degrade these models' performance thanks to their recurrent structure. 

In this paper, we address both of these deficiencies. We first provide a literature review of existing traffic prediction papers in Section \ref{sec:2_related_work}. From this literature review, we identify that state-of-the-art traffic prediction models rely on hybrid neural networks with recurrent structures. In addition, we also note the lack of existing literature that tackles long-term traffic prediction and propose our day-ahead traffic prediction problem in Section \ref{sec:3_problem_definition}. Short-term traffic prediction only needs short-term traffic pattern, but for long-term traffic, we require short, medium and long-term traffic patterns. We also describe in detail these challenges of long-term prediction compared to short-term in Section \ref{sec:3_problem_definition}.

Afterward, we describe our Transformer model called TrafFormer in Section \ref{sec:4_methodology}. Our TrafFormer model addresses the weaknesses of the commonly used recurrent models. In addition to the model itself, we add a time-day embedding, which captures the medium and long-term traffic pattern; and a spatio-temporal multi-head attention that captures the impact of differing time steps and traffic sensor locations. In Section \ref{sec:5_experiments}, we perform experiments comparing our model with several common deep neural network models as well as existing state-of-the-art traffic prediction models and show that our model provides the best overall performance. We also perform a set of ablation tests to explore the TrafFormer settings and our dataset-related design choices, which will help future traffic prediction research. Finally, we conclude our paper in Section \ref{sec:6_conclusion}.

To summarize, our contributions are:

\begin{itemize}
    \item We introduce the problem of long-term traffic prediction and discuss the differences and challenges when compared to the more conventional short-term traffic prediction.
    \item We formulate a novel Transformer model with time and day embedding, which we call TrafFormer. TrafFormer sidesteps the issue of using recurrent models, and our time and day embedding captures medium and long-term traffic patterns.
    \item We conduct a set of experiments comparing our model with state-of-the-art traffic prediction papers. These experiments show that our model overall performs the best. 
    \item We perform a set of ablation tests to show the importance of the different components of our model. 
\end{itemize}

\section{Related Work}\label{sec:2_related_work}

Traffic prediction is a well-studied field. Its research dates back to the late 70s, where classical statistical models were widely used. Out of these, the Autoregressive Integrated Moving Average (ARIMA) family of models is the most popular. Ahmed and Cook were the first researchers to apply ARIMA to traffic prediction \cite{Ahmed1979}. Other authors also applied different versions and improvements to ARIMA \cite{Lee1999}\cite{Williams2001}\cite{Williams2003}\cite{Kamarianakis2002}. Despite its popularity, classical statistical models are relatively weak. This is because they are simple linear models which assume that the traffic is stationary. Consequently, they frequently fail when handling the complex, nonlinear traffic data \cite{Wu2016}\cite{Karlaftis2011}\cite{Li2018}\cite{Dai2017}. 

Due to the aforementioned deficiencies of classical statistical models, researchers flocked to machine learning models. Out of the different machine learning models, neural network is the most commonly used. One of the first neural network applications in traffic flow prediction was by \citet{Dougherty1993}. Since then, various improvements to the neural network structure have been proposed \cite{Vlahogianni2005}\cite{Zheng2006}\cite{Zhong2005}\cite{Chan2012}. While machine learning models, and especially neural network, are more powerful compared to statistical models, they are very hard to train efficiently. Thus, machine learning models during the 2000s utilize shallow and simple structures, limiting their prediction power. However, the increasing computational power, as well as theoretical and software improvements in recent times had made increasingly complex neural network models feasible to train. Thus, in the middle of the 2010s, researchers started to apply deep neural network models for traffic prediction. 

Deep neural networks consist of complex neural network models with a large number of layers. \citet{Huang2014} and \citet{Yisheng2015} were among the first to apply deep neural networks to the task of traffic prediction; where they used a Deep Belief Network and Stacked Autoencoder respectively. As time progresses, traffic prediction models are becoming increasingly complex and often utilize many types of networks in a hybrid manner. Recurrent Neural Network (RNN), Convolutional Neural Network (CNN) and Feedforward Neural Network (FNN) are commonly used as components in these models. Amongst these three, RNN models--such as the Long Short-Term Memory (LSTM) and Gated Recurrent Unit (GRU)--are the most commonly used. This is because traffic data consists of a temporal sequence of traffic readings, which fits RNN's purpose as it has a recurrent structure that is naturally suited for this sequential data. RNNs have been used for traffic prediction in past works such as the ones by \citet{Ma2015}, \citet{Tian2015}, \citet{Fouladgar2017} and \citet{Kang2017}. 

RNNs are designed to take as input a sequence of values to output a single predicted value. For instance, in traffic prediction, the input can be the traffic speed from the past hour, and the output is the predicted traffic speed 15 minutes into the future. This renders RNNs ineffective for multiple future traffic predictions (i.e., 15, 30, 45, and 60 minutes). In this scenario, researchers use a modification of RNNs called the sequence autoencoder. Sequence autoencoder, which is also known by other names such as Seq2seq and encoder-decoder sequence-to-sequence, was pioneered by \citet{Cho2014} and \citet{sutskever2014sequence}. The idea is a combination of autoencoder and RNNs; the model takes as input a sequence and outputs another sequence as the output. This enables multiple future traffic predictions. For instance, the model can take the traffic speed from the past hour and outputs a sequence of future traffic readings up to one hour into the future. Thanks to this capability, sequence autoencoders have become one of the most common models used in traffic prediction; especially when used as a main component of a hybrid deep neural network, as in the works of \citet{He2019}, \citet{Do2019}, \citet{Pan2019} and \citet{Li2022Dynamic}. For a more in-depth discussion about these deep neural network models, we refer viewers to the survey paper \cite{Tedjopurnomo2022Survey}. 
\section{Problem Definition}\label{sec:3_problem_definition}

Traffic prediction concerns the usage of a learnable function that takes as input the historical traffic data from several previous time-steps in order to predict the traffic in the future. Two main types of traffic data used are traffic flow and traffic speed. Traffic flow is denoted as the total number of vehicles detected in a target detection site during a certain time period. Traffic speed is denoted as the average traveling speed of vehicles detected in a target detection site during a certain time period. We will use the general term ``traffic" to refer to both traffic flow and traffic speed. The traffic prediction problem can be denoted as:

\begin{equation}
	\hat{\trafficy}_{t+T'} = f([X_{t-T+1},X_{t-T},...,X_{t}])
\end{equation}

To predict the future traffic $\trafficy$ at time $t + T'$, we feed past traffic readings $X$ of up to $T$ time steps in the past into an arbitrary function $f$. The objective is to find the model parameters which minimize the error between the predicted traffic and the observed traffic:

\begin{equation}
	\theta^* = {\mathrm{arg\,}}\underset{\theta^*}{\mathrm{min}}\, L(\trafficy_{t+T'},\hat{\trafficy}_{t+T'};\theta^*)
\end{equation}

Where: 
\begin{itemize}
	\item $\trafficy_t$ : The observed traffic at time $t$
	\item $\hat{\trafficy}_{t}$ : The predicted traffic at time $t$
	\item $T$ : Input sequence length, i.e., how many time steps of past data are used as the input. 
	\item $T'$ : Prediction horizon, i.e., how many time steps in the future the prediction is for. 
	\item $f$ : An arbitrary function that calculates the traffic prediction based on the input data. 
	\item $L$ : Loss function, which is the function that calculates the quality of the prediction. 
	\item $\theta^*$ : The optimal set of parameters for the function $f$
\end{itemize}

All of $f$, $L$, and $\theta^*$ depend on the actual model used. The function $f$ constitutes the prediction model--e.g., the deep neural network models we mentioned in the previous section--and $\theta$ depends on the model used. The loss function $L$ is used to measure the prediction error, which in turn will be used to train the prediction model through backpropagation. Backpropagation computes the gradient of the loss with respect to the weights of the network. This gradient is propagated backward to adjust the parameters $\theta$ in each layer of the model. 

While traffic prediction is a well-studied topic, most of the traffic prediction research limit themselves to predicting traffic up to 60 minutes into the future. Long-term traffic prediction, in which traffic of up to 24 hours in the future is predicted, is rarely explored despite its usefulness in allowing proactive measures against traffic congestion. In terms of the problem formulation, short and long-term traffic prediction are almost identical. The only difference lies in the prediction horizon $T'$. Despite the similarity of the two problems, long-term traffic prediction is more challenging due to the instability of long-term traffic pattern that worsens error propagation. Such problems are not present in short-term traffic prediction, as traffic fluctuations in the short-term (e.g., two hours, covering one-hour past data to predict next-hour traffic) are less extreme. In addition, unlike short-term traffic prediction that only needs to learn traffic patterns from recent timestamps, long-term (i.e., one-day-ahead prediction) requires learning three traffic patterns:

\begin{itemize}
    \item \textbf{Short-term pattern}. Short-term traffic pattern relates to traffic in the past hour. Despite the day-to-day differences in traffic, hourly patterns still exist. For instance, peak hours in the morning and evening occur at around the same time every weekday. Therefore, data from the past hour may provide important information to predict traffic for the next day. 
    \item \textbf{Medium-term pattern}. Medium-term pattern relates to traffic in the past day. This is crucial for the model as it provides information of how traffic fluctuates on a daily basis, which will help next-day prediction.
    \item \textbf{Long-term pattern}. Long-term pattern relates to the different traffic patterns on different days. Small fluctuations in traffic patterns may exist on different days and especially on the weekends vs weekdays scenario. Thus, this pattern is also important. 
\end{itemize}

Due to these difficulties and additional factors to capture, long-term prediction is a challenging task. In the next section, we will describe in detail our proposed method to tackle these challenges. 
\section{Methodology}\label{sec:4_methodology}

While RNNs and its Sequence Autoencoder variants are commonly used in short-term traffic prediction due to their effectiveness, these models are not as proficient at handling long-term input data. This is because the instability of long-term traffic causes prediction errors, which are then propagated further into the network thanks to the recurrent structure. Moreover, in sequence autoencoders, the input traffic data is summarized into a vector representation which is fed to the decoder to output the predicted traffic. This means that the decoder does not have direct access to the input sequence anymore and can only rely on the lossy vector representation of the input. 

\begin{figure}
    \caption{A standard Transformer and our TrafFormer model}
    \begin{subfigure}[t!]{0.5\textwidth}
	\centering
	\includegraphics[scale=0.55]{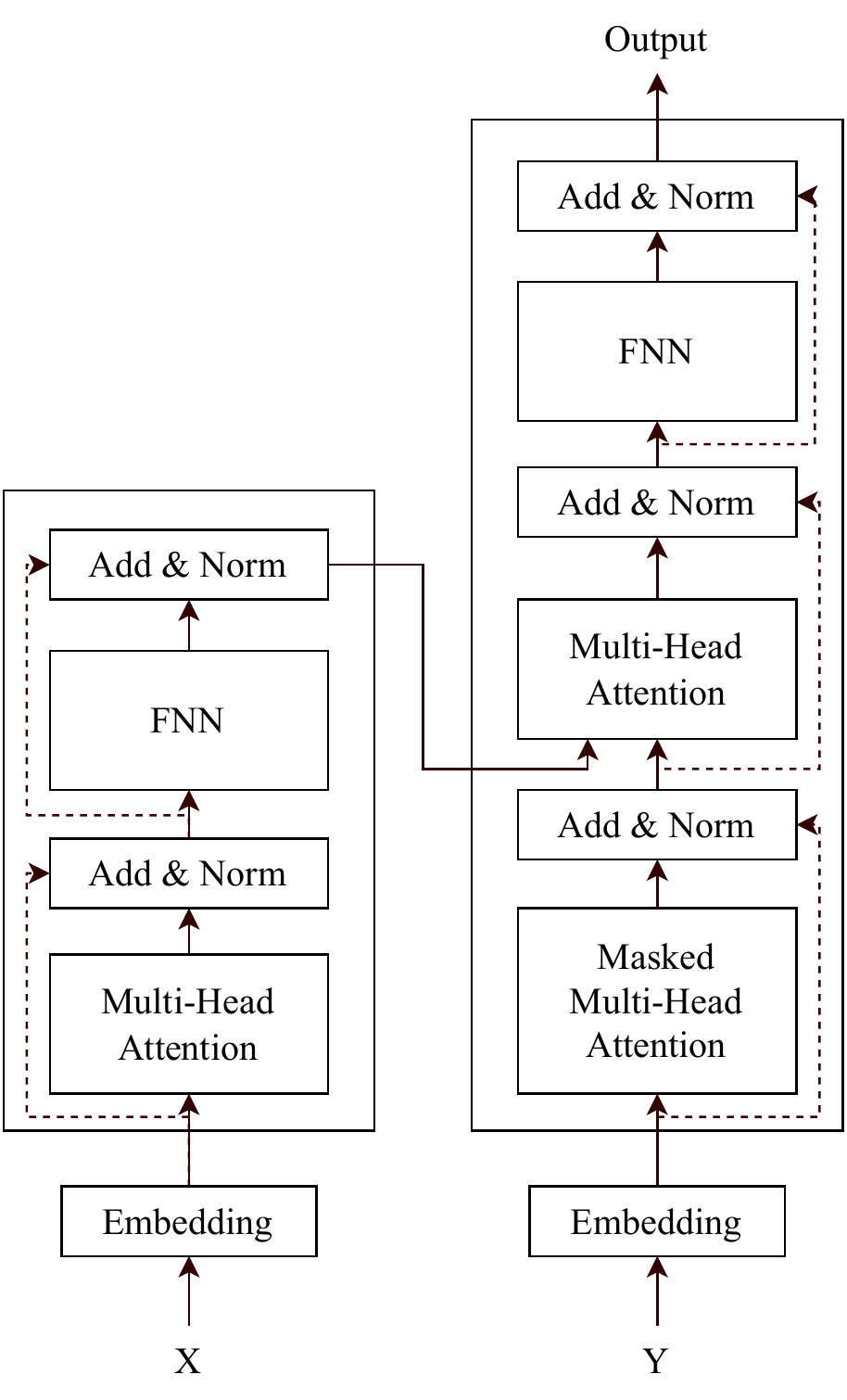}
	\caption{A transformer model}
	\label{ch2:fig:transformer}
    \end{subfigure} 
    \begin{subfigure}[t!]{0.5\textwidth}
	\centering
	\includegraphics[scale=0.6]{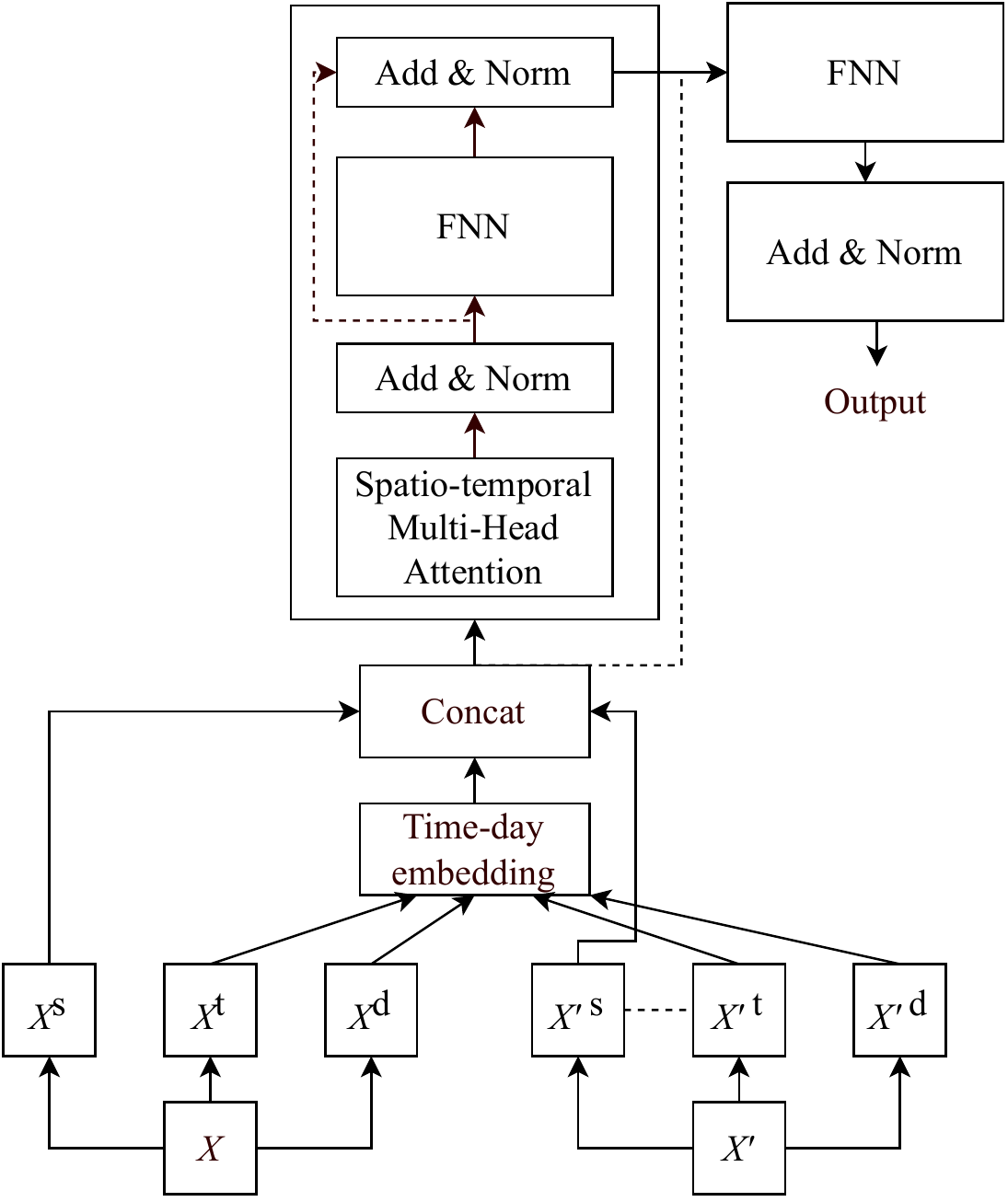}
	\caption{Our modified TrafFormer model}
	\label{ch3:fig:trafformer}
    \end{subfigure} 
\end{figure}

In order to sidestep these issues of long-term traffic prediction, we use a Transformer model. The Transformer model was proposed by \citet{vaswani2017attention} as an alternative to sequence autoencoders. Sequence autoencoders use RNNs as the encoder and decoder and thus process the data step-by-step through the recurrence process. On the other hand, Transformer uses an attention mechanism instead. This attention mechanism attends to every time step at once. We illustrate a Transformer in Figure \ref{ch2:fig:transformer}.

For ease of explanation, we will describe this model in the context of machine translation from English to French; as Transformers were originally used in NLP applications. The Transformer model can be divided into the encoder (left side) which takes $X$ as the input English sentence and the decoder (right side) which takes $Y$ as the French translation of the input English sentence. Multiple layers of encoders and decoders can be stacked to form a more complex and potentially more powerful model. Because the input sequence $X$ and output sequence $Y$ are in the form of sentences, each word first needs to be transformed into an embedding vector through the embedding layers. Both the encoder and decoder use the multi-head attention module. This multi-head attention gives Transformers the ability to attend to every time step at once. The encoder's Transformer enables every English word in the sentence to attend to any word in that sentence. For the French sentence, masked multi-head attention is required instead such that when the $n$-th English word is being predicted, the Transformer does not have knowledge of words after the $n$-th word. In addition, Transformers also use several addition and normalization layers, FNN layers, and residual connections denoted with the dashed arrows.  

Transformers work in a similar fashion for traffic prediction. The only major difference is that the embedding layer is not used. The reason we use Transformers is twofold:

\begin{itemize}
    \item Transformers do not use recurrent structures. They instead opt for attention layers, therefore sidestepping the long-term instability of models that use these recurrent structures.
    \item Unlike sequence autoencoders, where the decoder can only access the vector representation of the input sequence, Transformers' multi-head attention allows its decoder to have direct access to the input sequence and selectively pay more or less attention to relevant parts of the sequence.  
\end{itemize}

In this section, we will describe our methodology in detail. We use a modified Transformer model for our long-term traffic prediction work, which we call \textbf{TrafFormer}. To the best of our knowledge, TrafFormer is the first application of the Transformer model for traffic prediction. In addition to this, we also propose two novel additions: our time and day embedding, and the addition of medium and long-term past traffic reading. In addition, we also modify the multi-head attention to take into account the correlation between different time steps and traffic sensor locations; forming the spatio-temporal multi-head attention module. Our overall model structure is provided in Figure \ref{ch3:fig:trafformer}. We describe the different components of our model below. 

\noindent\textbf{Short and medium term data.} Most existing traffic prediction models take only short-term traffic data as input. This is suitable for short-term prediction, as the traffic condition in the near future is likely to be affected by traffic in the near past. This approach however is not sufficient for long-term traffic prediction, as short-term past traffic patterns may not reflect medium-term (e.g., 1 day) future data. To address this issue, we use two inputs to our model: the short-term traffic data from the past hour, denoted by $X$ and medium-term data from the past day, denoted by $X'$. 

\begin{figure}[t!]
	\centering
	\includegraphics[scale=0.55]{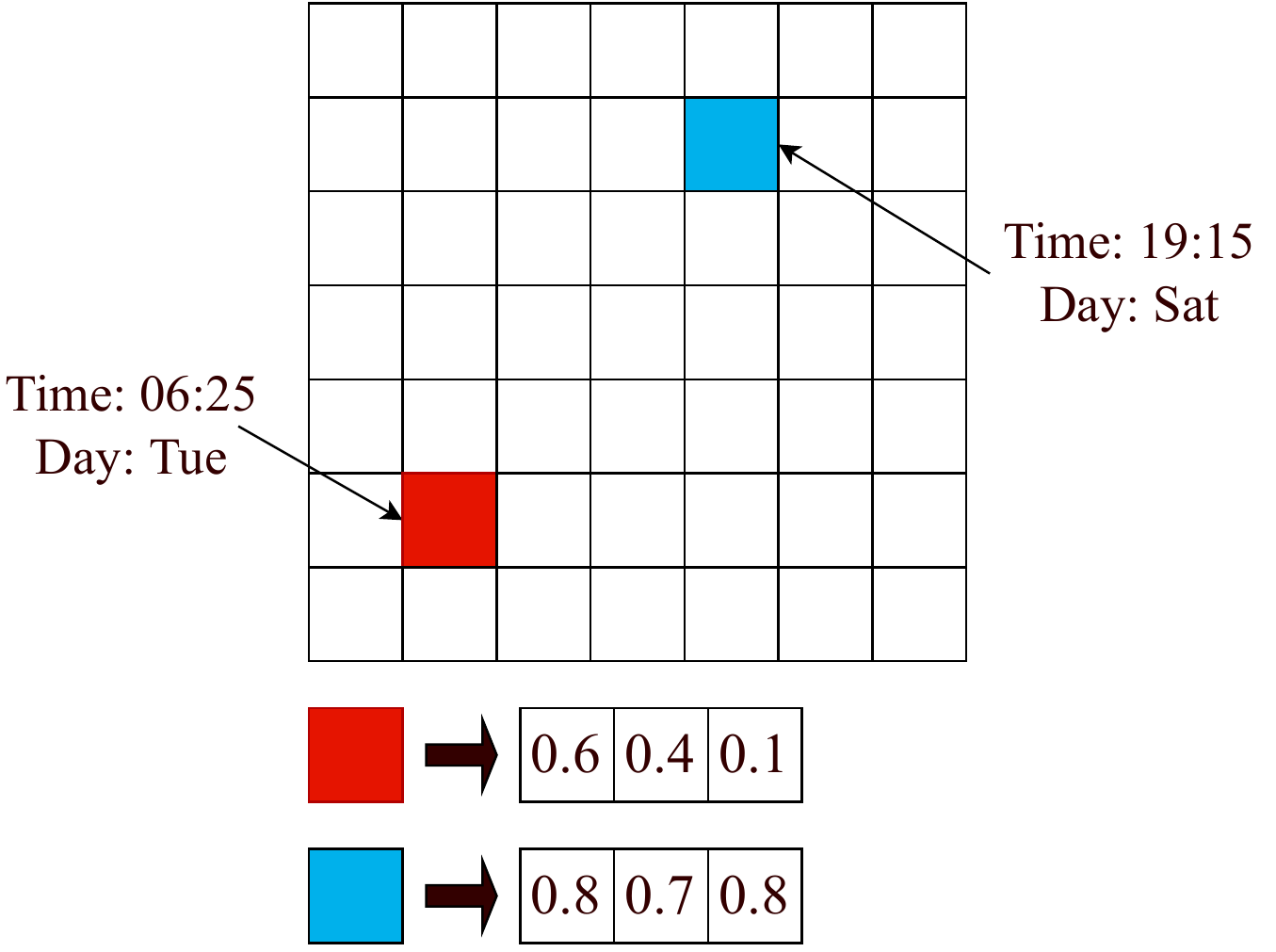}
	\caption{The time-day embedding process}
	\label{ch3:fig:timedayembed}
\end{figure}

\noindent\textbf{Time and day embedding.} The aforementioned short and medium term traffic data records how traffic changes over time in two different periods, but they do not have information of the time periods themselves. In order to incorporate this information, we use two types of embedding: time and day embedding. Specifically, for both the short and medium-term data, we record the time stamp and day, which we represent using the $t$ and $d$ superscript respectively in Figure \ref{ch3:fig:trafformer}; the $s$ subscript represents the past traffic reading. Afterwards, we transform them to different periods. For the time-in-day value, we use a 5 minute period window from 00:00-00:04 to 23:55-23:59, totaling 288 periods. For the day-in-week value, we use 7 for the different days. For each time period and day, we assign an embedding vector that captures the context of those periods. This is detailed in Figure \ref{ch3:fig:timedayembed} where two pairs of time and day recordings are used to perform lookups to retrieve their respective embedding vector of size 3. 

The embedding vectors are integrated into the TrafFormer model. Since the model is trained end-to-end, these vectors will also be trained, which will improve the effectiveness of the embeddings in representing the time-in-day and day-in-week information. Therefore, the embedding vectors help capture the medium-term and long-term patterns.

\noindent\textbf{TrafFormer Model.} After transforming the timestamp and day information for both the short and medium-term traffic data, we concatenate them together with the traffic readings and present them as an input to the Transformer part of our model. We use the encoder part of a Transformer and modified the multi-head attention such that it can attend to all time steps and all traffic sensors simultaneously, providing it the capability to capture both spatial and temporal patterns. In addition, we add a residual connection between the input and output of the Transformer encoder, denoted with the dashed line. 

In summary, we capture both the past-hour and past-day data for the short and medium-term patterns respectively. Then, we add the time and day embedding information to capture the medium and long-term patterns. Afterwards, we provide them to our Transformer model in order to capture both spatial and temporal patterns, utilizing the Transformer's power to simultaneously attend to all timestamps at once. 
\section{Experiments}\label{sec:5_experiments}

In this section, we will describe our experiment setup: our datasets, the hyperparameter settings of our models, and the baselines we compare our model with. We will then show our experiment results as well as several ablation tests that explore the importance of the different components of our TrafFormer model. 

\subsection{Datasets}

\begin{figure}
    \caption{Roads covered by the two datasets.}
    \begin{subfigure}[t!]{0.4\textwidth}
    	\centering
    	\includegraphics[scale=0.81]{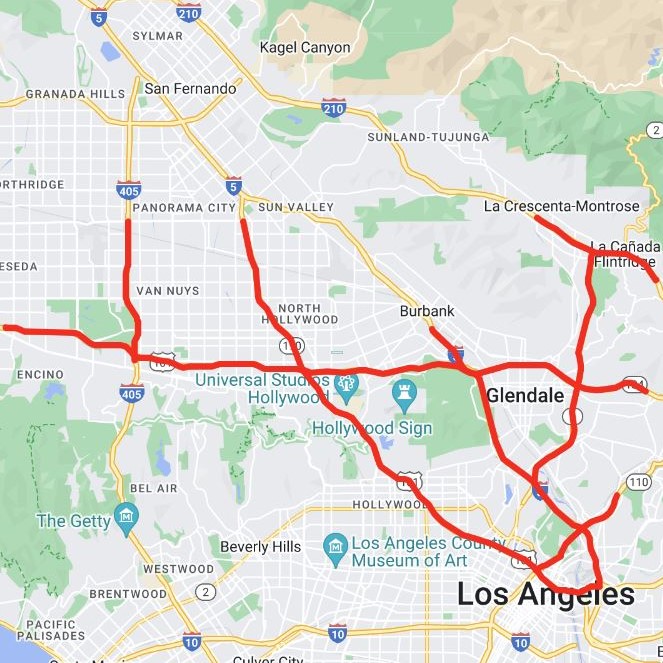}
    	\caption{METR-LA dataset}
    	\label{ch3:fig:dcrnn-metr}
    \end{subfigure} 
    \hfill
    \begin{subfigure}[t!]{0.4\textwidth}
    	\centering
    	\includegraphics[scale=0.585]{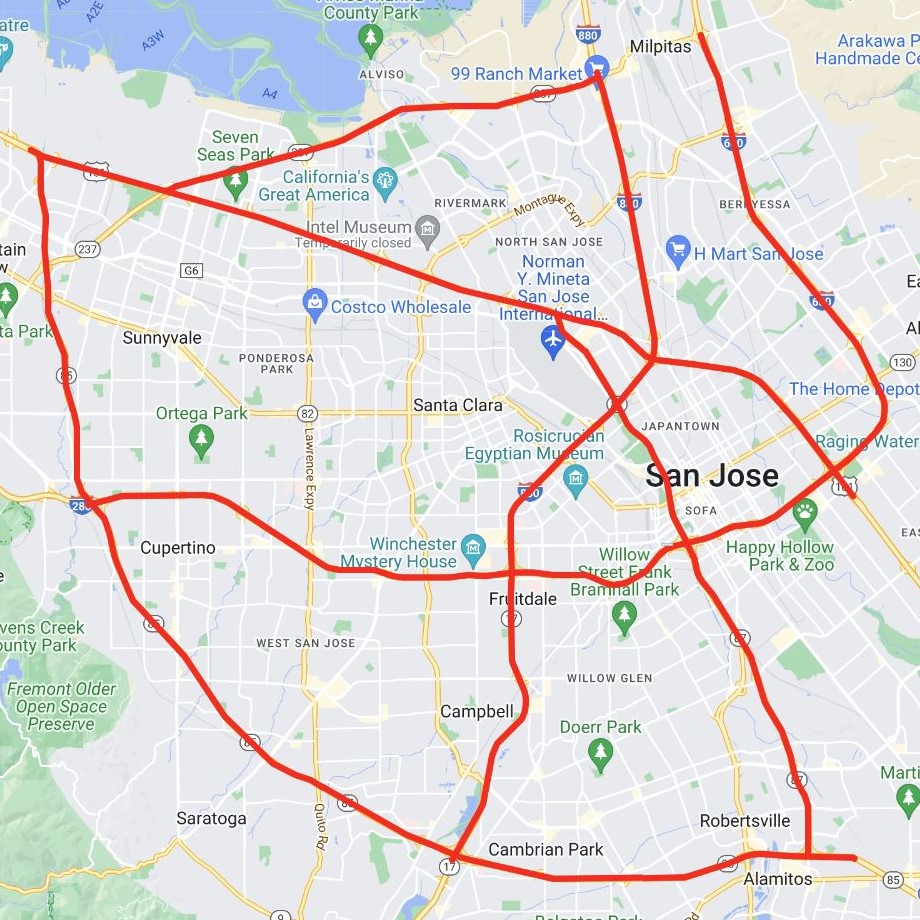}
    	\caption{PEMS-BAY dataset}
    	\label{ch3:fig:dcrnn-pems}
    \end{subfigure} 
\end{figure}

We use the following two dataset in our experiments, with both of them recording traffic speed: 

\begin{itemize}
    \item \textbf{METR-LA}\cite{Jagadish2014Big}. This dataset is collected from 207 traffic loop detectors on the highways of Los Angeles County from March 2012 to June 2012. The granularity of the recording is 5 minutes, resulting in 34,272 total speed readings for each traffic detector. We use this dataset due to its common usage in traffic prediction studies \cite{Li2017}\cite{Pan2019}\cite{Chen2019Gated}\cite{Chen2020Dynamic}\cite{Wang2020Traffic}\cite{Huang2021LTPHM}\cite{Li2022Dynamic}. The covered area is visualized in Figure \ref{ch3:fig:dcrnn-metr}. 
    \item \textbf{PEMS-BAY}\cite{Li2017}. This is a subset of the commonly-used Caltrans PEMS dataset. It covers 325 traffic loop detectors on the San Francisco Bay area from January 2017 to May 2017. The granularity of the recording is 5 minutes, resulting in 52,116 total speed readings for each traffic detector. This is also a commonly used dataset \cite{Li2017}\cite{Chen2019Gated}\cite{Wang2020Traffic}\cite{Huang2021LTPHM}\cite{Li2022Dynamic}. The covered area is visualized in Figure \ref{ch3:fig:dcrnn-pems}.
\end{itemize} 

We use these datasets mainly for reproducibility purposes, as these two datasets are some of the most commonly used ones in the literature. In addition, they cover a large area, have decent temporal coverage, and have different traffic patterns, as can be seen by the different experimental results we will present later in this section. 

\subsection{Model Settings}

For our TrafFormer model, as well as the baselines, we divide the dataset into training, validation, and test sets using a 70:10:20 ratio without shuffling; meaning that the first (in terms of the time period) 70\% of the data is assigned to the training set, the next 10\% into the validation set, and the final 20\% into the test set. This follows the setup used in \cite{Li2017} where these two datasets were also used. For the past hour data $X$, we use a granularity of 5 minutes, resulting in 12 time steps of past hour data. We use this setting as it is the default setting of the Caltrans dataset, the most commonly used granularity in the literature, and a fine enough granularity to capture short-term traffic patterns. For the past day data $X'$ and the ground truth value for the next day data $Y$, however, we cannot use this granularity as this will result in 288 time steps, which are prohibitively expensive. Therefore, instead of using the 5-minute granularity, we use a 2-hour granularity, which also results in 12 time steps of past and next-day data for $X'$ and $Y$ respectively. 

We use a sliding window approach to retrieve the dataset. Specifically, at a time-step $t$, we retrieve the traffic speed data $S$ using three sliding windows for $X$, $X'$ and \scalebox{2.2}{$\scriptscriptstyle y$} for the periods specified below:

\begin{itemize}
    \item $X  = [S_{t-11},S_{t-10},...,S_{t-1},S_{t}]$
    \item $X' = [S_{t-264},S_{t-240},...,S_{t-24},S_{t}]$
    \item $\scalebox{2.2}{$\scriptscriptstyle y$}  = [S_{t+24},S_{t+48},...,S_{t+264},S_{t+288}]$
\end{itemize}

We slide these three windows starting from $S_{264}$ to $S_{|S|-288}$ one timestep each time (i.e., 5 minute increments), where $|S|$ denotes the total timesteps for the entire dataset. The increment of 24 used in $X'$ and \scalebox{2.2}{$\scriptscriptstyle y$} denote the two hour period for the past and next day data. 

For the model structure, we set the time-day embedding matrix to be $288 \times 7$ in size, representing the time-in-day on 5 minute increments and day-in-week respectively. The size of the day and time embedding vectors is set to 64. All FNNs in the TrafFormer model have 64 computational nodes. We apply layer normalization for the traffic speed, which transforms the traffic speed reading such that the mean is 0 and the standard deviation is 1. We train our model using cyclical learning rate \cite{Smith2015Cyclical} with the \textit{`triangular2'} setting with step size equal to $2\times$ the number of minibatches in the dataset. The model is trained for a maximum of 50 epochs with early stopping if the validation loss does not decrease for 2 consecutive epochs. In practice, we find that the model training ends at around the 6-12 epochs mark. We train the model using the Mean Absolute Error (MAE) loss. The code for our model is publicly available\footnote{https://github.com/david-tedjopurnomo/TrafFormer}.
\subsection{Baselines}

We compare our methods against three general deep neural network structures and three state-of-the-art benchmarks; the latter three benchmarks were implemented only for short-term prediction. We describe these models below:  

\begin{itemize}
    \item \textbf{FNN}. We replace the transformer encoder part of TrafFormer with a 3-layer FNN with 256, 64, and 16 nodes respectively. Similar to TrafFormer, this model receives both the past-day and past-hour data as the input. 
    \item \textbf{Stacked GRU\cite{Cho2014}}. We use a stack of 2 GRUs with 128 hidden units each. Similar to TrafFormer, this model receives both the past-day and past-hour data as the input. 
    \item \textbf{LSTM Sequence Autoencoder\cite{sutskever2014sequence}}. We adapt the sequence autoencoder structure with 2 LSTMs for the encoder and decoder, each with 64 hidden nodes. Similar to TrafFormer, this model receives both the past-day and past-hour data as the input. 
    \item \textbf{Diffusion Convolution Recurrent Neural Network (DCRNN)\cite{Li2017}}. The authors proposed a GRU sequence autoencoder with a graph diffusion convolution in order to capture the temporal dependencies of traffic data as well as spatial dependencies between different traffic detectors. We follow the default architecture and only feed the past hour's data for the next-day prediction. 
    \item \textbf{Spatio-Temporal Graph Convolutional Networks (STGCN)\cite{Yu2018}}. The authors proposed a novel spatio-temporal convolution block consisting of a Graph CNN to capture the structure of the road network and a gated CNN to extract temporal features. Both of these components are then fused to give the model the capability to capture both the spatial and temporal features simultaneously. We follow the default architecture and only feed the past hour's data for the next-day prediction. 
    \item \textbf{Graph and Attentive Multi-path Convolutional Network (GAMCN)\cite{qi2022gamcn}}. The authors proposed a novel LPGCN module that captures the spatial correlation between traffic conditions in close and far-away locations alike, and a temporal correlation modeling with multi-path temporal convolution. These two components are fused with a gated fusion layer. We follow the default architecture and only feed the past hour's data for the next-day prediction. 
\end{itemize}

We evaluate all models using the Root Mean Squared Error (RMSE), Mean Absolute Error (MAE), and Mean Absolute Percentage Error (MAPE) values formulated below:

\begin{equation}
	RMSE(\trafficy, \hat{\trafficy}) =  \sqrt{\frac{\sum_{i=1}^{n} (\trafficy_i-\hat{\trafficy_i})^2}{n}}
\end{equation}

\begin{equation}
	MAE(\trafficy, \hat{\trafficy}) =  \frac{\sum_{i=1}^{n} \left|\trafficy_i-\hat{\trafficy_i}\right|}{n}
\end{equation}

\begin{equation}
	MAPE(\trafficy, \hat{\trafficy}) =  \frac{1}{n}\sum_{i=1}^{n}\left|\frac{\trafficy_i-\hat{\trafficy_i}}{\trafficy_i}\right|
\end{equation}

Where $\trafficy$ is the actual traffic speed $\hat{\trafficy}$ is the predicted traffic speed and $n$ is the number of samples. For all metrics, a lower value denotes better performance. We show the experiment results not only for day-ahead prediction, but also for 6, 12 and 18 hours future prediction in order to show the performance of all models on different prediction horizons. 

All methods are run on a Red Hat Enterprise Linux server with an Intel Xeon E5-2690 CPU, 52 GB of RAM, and two Nvidia Tesla P100 GPUs with 16 GB of memory. 

\subsection{Experiment Results}

The experiment results for the METR-LA and PEMS-BAY dataset are provided in Tables \ref{ch3:tab:exp_metr} and \ref{ch3:tab:exp_pems} respectively. For all models, the performance on the PEMS-BAY dataset is better compared to the METR-LA dataset. This is due to two factors:

\begin{enumerate}
    \item METR-LA contains a lot of missing data; out of the 1,396,008 data points, 175,677 are missing, which is approximately 12.6\%. In contrast, PEMS-BAY contains 3,351,726 data points, but only 217 of them are missing. 
    \item As can be seen in Figure \ref{ch3:fig:dcrnn-metr}, the METR-LA dataset spans both urban and rural areas; it covers major tourism areas such as Hollywood, and less-populated areas in the West--which is located just north of California's State Parks--and northeast--which nears the mountainous areas of Southern California. On the other hand, the PEMS-BAY dataset, shown in Figure \ref{ch3:fig:dcrnn-pems} covers mostly the metropolitan areas of Silicon Valley and San Francisco Bay. Thus, the PEMS-BAY dataset is more homogeneous with relatively smaller traffic speed fluctuations within that area, contributing to its lower prediction error. 
\end{enumerate}

\begin{table}[!t]
\caption{Experiment results for the METR-LA dataset.}
\label{ch3:tab:exp_metr}
\resizebox{\columnwidth}{!}{%
\begin{tabular}{@{}cccclccclccclccc@{}}
\toprule
\multirow{2}{*}{Method} & \multicolumn{3}{c}{6 hours}                      &  & \multicolumn{3}{c}{12 hours}                     &  & \multicolumn{3}{c}{18 hours}                     &  & \multicolumn{3}{c}{24 hours}                     \\ \cmidrule(l){2-16} 
                        & RMSE          & MAE           & MAPE             &  & RMSE          & MAE           & MAPE             &  & RMSE          & MAE           & MAPE             &  & RMSE          & MAE           & MAPE             \\ \midrule
FNN                     & 11.46         & 6.05          & 22.14\%          &  & 11.52         & 6.08          & 22.46\%          &  & 11.54         & 6.10          & 22.56\%          &  & 11.46         & 6.09          & 22.36\%          \\
Stacked GRU             & 10.70         & 5.68          & 19.72\%          &  & 10.78         & 5.72          & 20.03\%          &  & 10.77         & 5.73          & 20.05\%          &  & 10.69         & 5.71          & 19.78\%          \\
Seq2Seq LSTM            & 9.68          & 5.02          & 16.83\%          &  & 9.74          & 5.06          & 17.09\%          &  & 9.77          & 5.07          & 17.17\%          &  & 9.72          & 5.06          & 17.05\%          \\
DCRNN                   & 8.99          & 4.29          & \textbf{13.20\%} &  & 9.33          & 4.48          & 13.61\%          &  & 9.66          & 4.64          & 13.98\%          &  & 9.63          & 4.66          & 14.29\%          \\
STGCN                   & 14.86         & 8.76          & 21.07\%          &  & 15.88         & 9.31          & 22.24\%          &  & 15.31         & 9.12          & 21.99\%          &  & 13.60         & 7.83          & 21.29\%          \\
GAMCN                   & 13.33         & 6.30          & 16.23\%          &  & 12.52         & 5.85          & 15.90\%          &  & 11.51         & 5.35          & 15.07\%          &  & 9.61          & 4.60          & 13.84\%          \\
TrafFormer              & \textbf{8.47} & \textbf{4.11} & 13.25\%          &  & \textbf{8.53} & \textbf{4.14} & \textbf{13.39\%} &  & \textbf{8.56} & \textbf{4.16} & \textbf{13.50\%} &  & \textbf{8.59} & \textbf{4.18} & \textbf{13.57\%} \\ \bottomrule
\end{tabular}
}
\end{table}
\begin{table}[!t]
\caption{Experiment results for the PEMS-BAY dataset.}
\label{ch3:tab:exp_pems}
\resizebox{\columnwidth}{!}{%
\begin{tabular}{@{}clllllllllllllll@{}}
\toprule
\multirow{2}{*}{Method} & \multicolumn{3}{c}{6 hours}                                                   &           & \multicolumn{3}{c}{12 hours}                                                  &           & \multicolumn{3}{c}{18 hours}                                                  &           & \multicolumn{3}{c}{24 hours}                                                  \\ \cmidrule(l){2-16} 
                        & \multicolumn{1}{c}{RMSE} & \multicolumn{1}{c}{MAE} & \multicolumn{1}{c}{MAPE} &           & \multicolumn{1}{c}{RMSE} & \multicolumn{1}{c}{MAE} & \multicolumn{1}{c}{MAPE} &           & \multicolumn{1}{c}{RMSE} & \multicolumn{1}{c}{MAE} & \multicolumn{1}{c}{MAPE} &           & \multicolumn{1}{c}{RMSE} & \multicolumn{1}{c}{MAE} & \multicolumn{1}{c}{MAPE} \\ \midrule
FNN                     & 8.27                     & 3.83                    & 11.21\%                  &           & 8.25                     & 3.83                    & 11.17\%                  &           & 8.25                     & 3.83                    & 11.17\%                  &           & 8.23                     & 3.84                    & 11.12\%                  \\
Stacked GRU             & 6.20                     & 2.82                    & 7.17\%                   &           & 6.20                     & 2.84                    & 7.19\%                   &           & 6.20                     & 2.84                    & 7.19\%                   &           & 6.20                     & 2.86                    & 7.20\%                   \\
Seq2Seq LSTM            & 6.22                     & 2.83                    & 7.13\%                   &           & 6.22                     & 2.83                    & 7.11\%                   &           & 6.22                     & 2.83                    & 7.11\%                   &           & 6.24                     & 2.85                    & 7.12\%                   \\
DCRNN                   & 5.54                     & 2.45                    & 6.07\%                   &           & 5.83                     & 2.54                    & 6.43\%                   &           & 5.83                     & 2.54                    & 6.43\%                   &           & 5.84                     & 2.58                    & 6.32\%                   \\
STGCN                   & 5.76                     & 2.65                    & 6.30\%                   &           & 6.18                     & 2.83                    & 6.86\%                   &           & 6.18                     & 2.83                    & 6.86\%                   &           & 6.81                     & 3.07                    & 7.80\%                      \\
GAMCN                   & \textbf{5.16}            & \textbf{2.30}           & \textbf{5.56\%}          &           & \textbf{5.18}            & \textbf{2.31}           & \textbf{5.59\%}          &           & \textbf{5.18}            & \textbf{2.31}           & \textbf{5.59\%}          &           & \textbf{5.22}            & \textbf{2.34}           & \textbf{5.63\%}          \\
TrafFormer              & 5.47                     & 2.59                    & 6.15\%                   & \textbf{} & 5.43                     & 2.58                    & 6.09\%                   & \textbf{} & 5.43                     & 2.58                    & 6.09\%                   & \textbf{} & 5.46                     & 2.59                    & 6.14\%                   \\ \bottomrule
\end{tabular}
}
\end{table}

From the METR-LA results, we can see that TrafFormer manages to outperform all models on all metrics and all time steps except for the 6-hours MAPE result where TrafFormer takes a very close second place to DCRNN. Moreover, we observe that TrafFormer is the most resilient to increasing future prediction horizon, as can be seen by the smaller performance degradation between the 6 hours and 24 hours results. DCRNN is the second-best performer, followed by Seq2seq LSTM. These three models are the only ones capable of achieving sub-10 RMSE results on all prediction horizon. We also notice that our model, as well as the three general deep neural network models--FNN, Stacked GRU and Seq2seq LSTM--have better resilience across all prediction horizon compared to the three state-of-the-art baselines. 

From the PEMS-BAY results, we observe that GAMCN performs the best. TrafFormer competes with DCRNN, beating DCRNN on all prediction horizons in terms of RMSE, losing in terms of MAE, and generally outperforming DCRNN in terms of MAPE. The three general deep neural network models perform significantly worse, with none of them managing to achieve a sub-5 RMSE value. Contrary to the METR-LA experiments, the results from the different prediction horizon are relatively consistent for all models except STGCN. This consistency is due to PEMS-BAY's smaller fluctuations. While GAMCN outperforms TrafFormer in PEMS-BAY, it does not perform well on the METR-LA dataset--especially for the 6, 12 and 18 hours experiment. Thus, from these results, TrafFormer is the overall best performer. 

\begin{figure}
    \centering
    \includegraphics[width=1.0\textwidth]{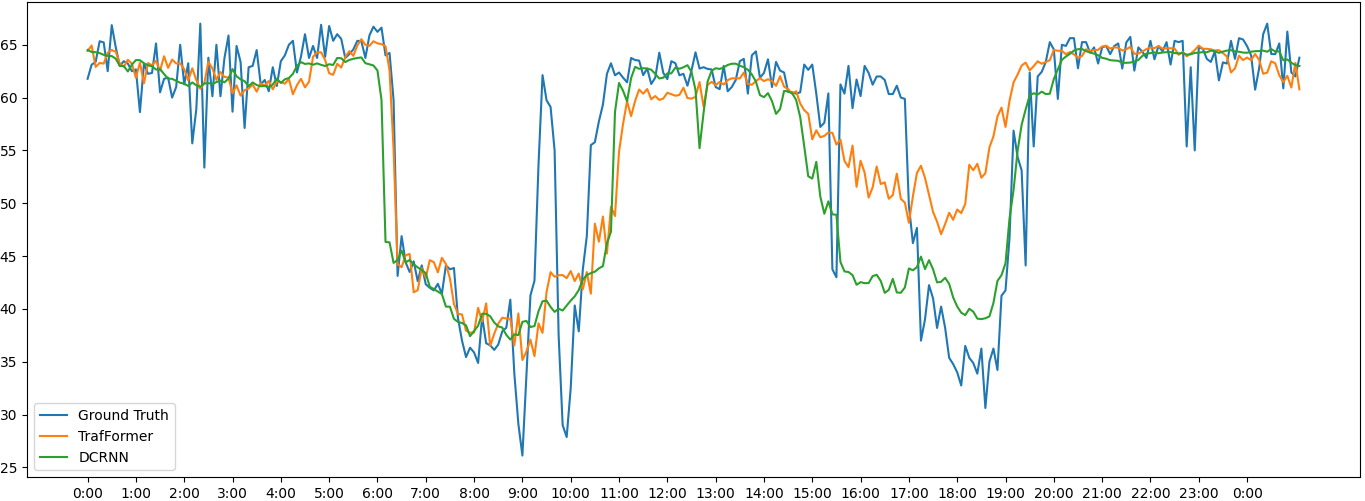}
    \caption{Prediction comparison between TrafFormer and DCRNN for the METR-LA dataset.}
    \label{ch3:fig:speedplot}
\end{figure}

In Figure \ref{ch3:fig:speedplot}, we show the METR-LA speed prediction of TrafFormer compared to DCRNN. This data is taken on 1 March 2012 from detector ID ``772140" which is located at Ventura Highway. We only choose DCRNN because it is the most competitive state-of-the-art baseline in the METR-LA experiments. From this figure, we observe that TrafFormer is better at identifying the start of a rush hour where traffic speed slows down. This can be seen from how DCRNN identifies the start of the speed drops at 06:00 and 14:30 too early. In addition, compared to DCRNN, TrafFormer has a tendency to underfit the data compared to DCRNN during periods of unstable traffic speed. This can be seen in the 15:00-19:00 stretch. At around 15:15, there is a significant, short-term speed drop. DCRNN is affected by this drop, which causes its prediction to be negatively affected until 17:00 whereas TrafFormer is not as affected. The same underfitting scenario is seen again during 17:00-19:00, but this time it affects TrafFormer negatively as it did not adapt to the traffic speed drop as good as DCRNN.

\begin{figure}
    \centering
    \includegraphics[width=1.0\textwidth]{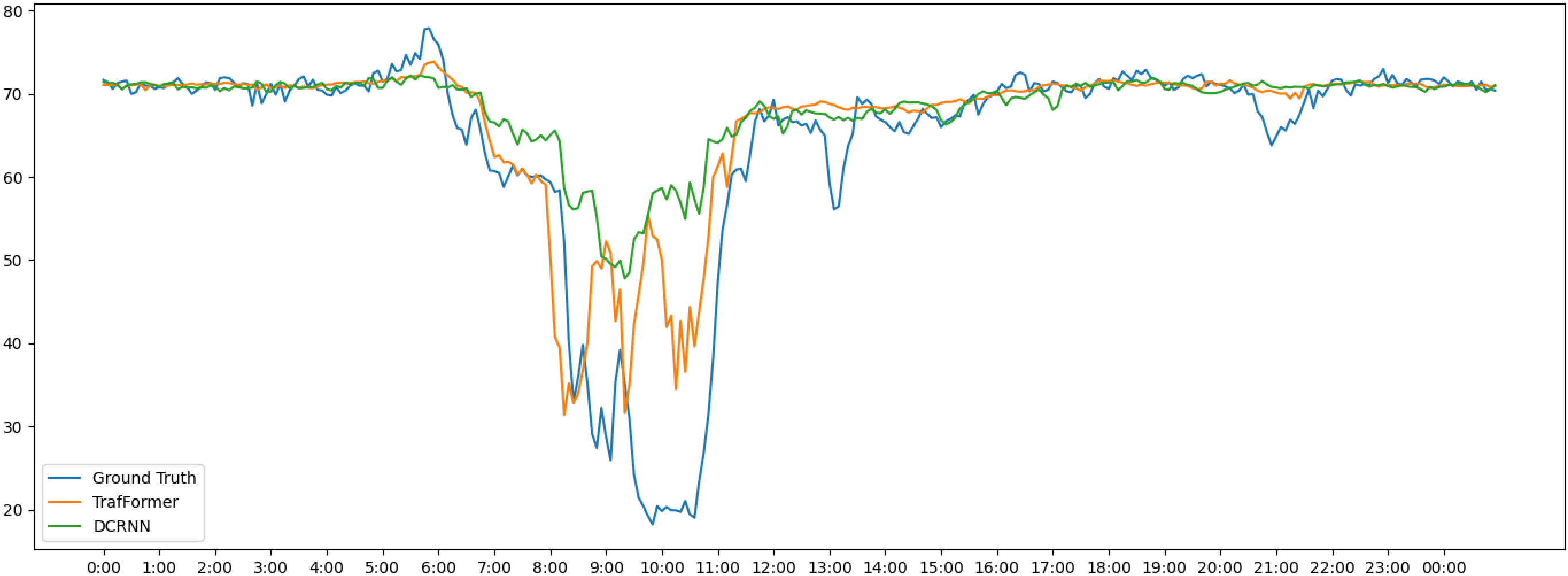}
    \caption{Prediction comparison between TrafFormer and DCRNN for detector ID ``400001" in the PEMS-BAY dataset.}
    \label{ch3:fig:speedplot-400001}
\end{figure}
\begin{figure}
    \centering
    \includegraphics[width=1.0\textwidth]{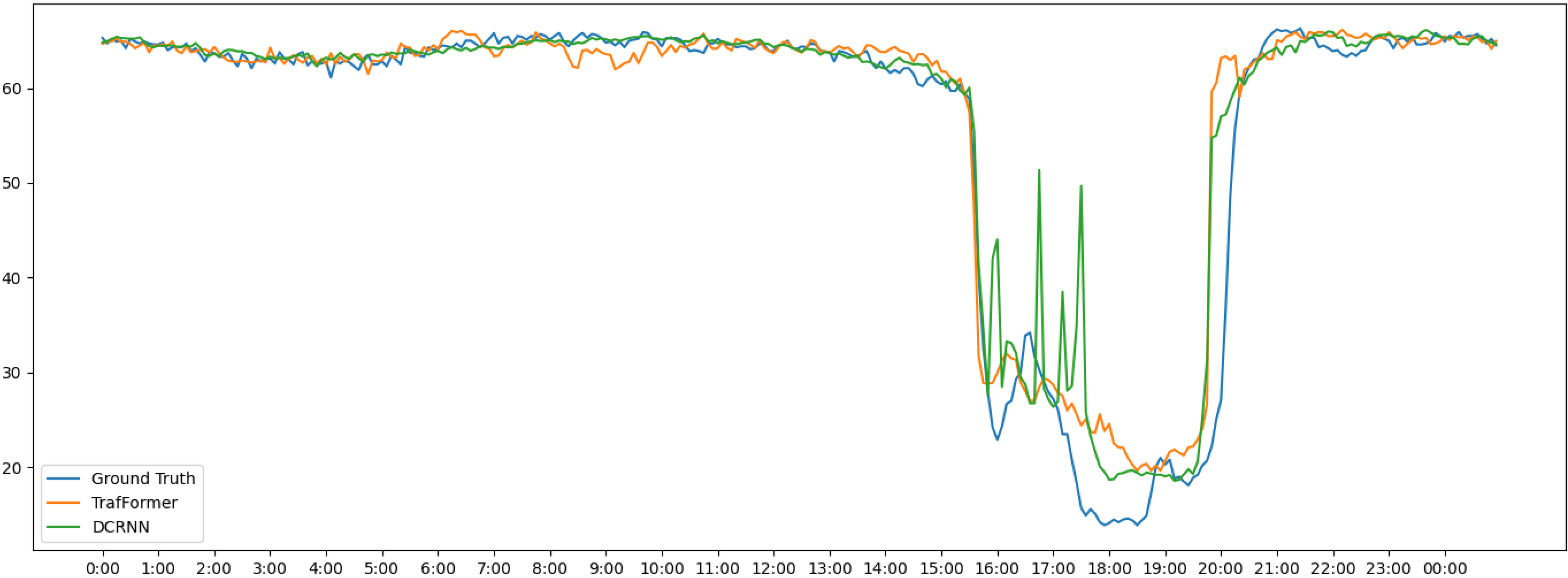}
    \caption{Prediction comparison between TrafFormer and DCRNN for detector ID ``400001" in the PEMS-BAY dataset.}
    \label{ch3:fig:speedplot-400017}
\end{figure}

In Figures \ref{ch3:fig:speedplot-400001} and \ref{ch3:fig:speedplot-400017}, we compare the traffic speed prediction between TrafFormer and DCRNN on two sites in the PEMS-BAY dataset. These comparisons are taken on 1 January 2017 from detectors ``400001" and ``400017", which are located east of Norman Y. Mineta San Jose International Airport, and at Highway 85 north of the Good Samaritan Hospital, respectively. From these two results, we can see that just like in the METR-LA comparison, TrafFormer performs better than DCRNN, but only slightly as demonstrated in the results in Figure \ref{ch3:tab:exp_pems} as well. One point to note is the different traffic speed drop patterns between the two, as well as the METR-LA dataset. In METR-LA, both morning and evening rush hours affected the traffic speed. In PEMS-BAY, however, we see that detector ``400001" only suffers from speed drops in the morning rush hours while detector ``400017" only suffers from speed drops in the evening rush hours. This shows that different detection sites can have vastly different traffic pattern, which makes it important to learn the spatial correlation. 

\subsection{Ablation Tests}

\begin{table}[t!]
\caption{Ablation test results for the METR-LA dataset.}
\label{ch3:tab:abl_metr}
\resizebox{\columnwidth}{!}{%
\begin{tabular}{llllllllllllllll}
\hline
\multicolumn{1}{c}{\multirow{2}{*}{Method}} & \multicolumn{3}{c}{6 hours}                                                   &  & \multicolumn{3}{c}{12 hours}                                                  &  & \multicolumn{3}{c}{18 hours}                                                  &  & \multicolumn{3}{c}{24 hours}                                                  \\ \cline{2-16} 
\multicolumn{1}{c}{}                        & \multicolumn{1}{c}{RMSE} & \multicolumn{1}{c}{MAE} & \multicolumn{1}{c}{MAPE} &  & \multicolumn{1}{c}{RMSE} & \multicolumn{1}{c}{MAE} & \multicolumn{1}{c}{MAPE} &  & \multicolumn{1}{c}{RMSE} & \multicolumn{1}{c}{MAE} & \multicolumn{1}{c}{MAPE} &  & \multicolumn{1}{c}{RMSE} & \multicolumn{1}{c}{MAE} & \multicolumn{1}{c}{MAPE} \\ \hline
Default                                     & 8.47                     & \textbf{4.11}           & \textbf{13.25\%}         &  & 8.53                     & \textbf{4.14}           & \textbf{13.39\%}         &  & 8.56                     & \textbf{4.16}           & \textbf{13.50\%}         &  & 8.59                     & \textbf{4.18}           & \textbf{13.57\%}         \\
MSE                                         & \textbf{8.03}            & 4.59                    & 14.05\%                  &  & \textbf{8.17}            & 4.66                    & 14.38\%                  &  & \textbf{8.22}            & 4.7                     & 14.47\%                  &  & \textbf{8.14}            & 4.67                    & 14.19\%                  \\
Hour only                                   & 9.54                     & 4.81                    & 15.61\%                  &  & 9.64                     & 4.87                    & 15.95\%                  &  & 9.68                     & 4.89                    & 16.07\%                  &  & 9.63                     & 4.88                    & 15.97\%                  \\
Day only                                    & 12.01                    & 5.91                    & 22.59\%                  &  & 12.16                    & 5.99                    & 23.06\%                  &  & 12.25                    & 6.05                    & 23.33\%                  &  & 12.14                    & 6.01                    & 23.05\%                  \\
Cyclical                               & 11.18                    & 6.02                    & 21.04\%                  &  & 11.26                    & 6.06                    & 21.35\%                  &  & 11.31                    & 6.09                    & 21.49\%                  &  & 11.25                    & 6.07                    & 21.32\%                  \\
Speed only                                  & 11.33                    & 6.00                    & 20.84\%                  &  & 11.30                    & 5.98                    & 20.87\%                  &  & 11.49                    & 6.13                    & 21.44\%                  &  & 11.39                    & 6.06                    & 21.15\%                  \\ \hline
\end{tabular}
}
\end{table}
\begin{table}[t!]
\caption{Ablation test results for the PEMS-BAY dataset.}
\label{ch3:tab:abl_pems}
\resizebox{\columnwidth}{!}{%
\begin{tabular}{lccccccccccccccc}
\hline
\multicolumn{1}{c}{\multirow{2}{*}{Method}} & \multicolumn{3}{c}{6 hours}                     & \multicolumn{1}{l}{} & \multicolumn{3}{c}{12 hours}                    & \multicolumn{1}{l}{} & \multicolumn{3}{c}{18 hours}                    & \multicolumn{1}{l}{} & \multicolumn{3}{c}{24 hours}                    \\ \cline{2-16} 
\multicolumn{1}{c}{}                        & RMSE          & MAE           & MAPE            & \multicolumn{1}{l}{} & RMSE          & MAE           & MAPE            & \multicolumn{1}{l}{} & RMSE          & MAE           & MAPE            & \multicolumn{1}{l}{} & RMSE          & MAE           & MAPE            \\ \hline
Default                                     & 5.47          & \textbf{2.59} & \textbf{6.15\%} &                      & 5.43          & \textbf{2.58} & \textbf{6.09\%} &                      & 5.43          & \textbf{2.57} & \textbf{6.08\%} &                      & 5.46          & \textbf{2.59} & \textbf{6.14\%} \\
MSE                                         & \textbf{5.31} & 2.76          & 6.47\%          &                      & \textbf{5.32} & 2.78          & 6.46\%          &                      & \textbf{5.31} & 2.77          & 6.42\%          &                      & \textbf{5.37} & 2.81          & 6.50\%          \\
Hour only                                   & 6.21          & 2.85          & 7.15\%          &                      & 6.22          & 2.85          & 7.12\%          &                      & 6.24          & 2.86          & 7.13\%          &                      & 6.27          & 2.88          & 7.19\%          \\
Day only                                    & 6.21          & 3.13          & 7.41\%          &                      & 6.19          & 3.13          & 7.46\%          &                      & 6.18          & 3.12          & 7.43\%          &                      & 6.23          & 3.16          & 7.49\%          \\
Cyclical                               & 8.51          & 4.28          & 11.11\%         &                      & 8.54          & 4.29          & 11.09\%         &                      & 8.56          & 4.31          & 11.11\%         &                      & 8.61          & 4.34          & 11.21\%         \\
Speed only                                  & 8.57          & 4.26          & 11.24\%         &                      & 8.50          & 4.27          & 11.03\%         &                      & 8.66          & 4.35          & 11.31\%         &                      & 8.48          & 4.27          & 10.94\%         \\ \hline
\end{tabular}
}
\end{table}

We perform several ablation tests to understand the importance of the components of our TrafFormer model as well as the different aspects of the input dataset. We perform five sets of experiments:

\noindent\textbf{MSE}. In this test, we replaced the loss function in the training from MAE to MSE; which is the Mean Squared Error, i.e. RMSE without the square root operation. The results show noticeable improvements of the RMSE results at the cost of worse MAE and MAPE score. Since RMSE and MSE take the square of the error between the actual and predicted value, they greatly penalize severely wrong predictions. On the other hand, MAE and MAPE only records the absolute difference of the prediction and therefore do not incur extra penalties for severely wrong predictions. The choice between RMSE or MSE, and MAE or MAPE depends on the task. On tasks where large errors are undesirable, RMSE is the better choice. On the other hand, on tasks where this is not the case and a lower overall error is to be prioritized, MAE and MAPE could be chosen instead. The choice between MAE and MAPE also depends on the context, as MAPE records the error relative to the ground truth value as percentages whereas MAE simply records the error as it is. 

\noindent\textbf{Hour only}. Here we only use the past-hour data $X$ as input; i.e., the past-day data $X'$ is removed. In this setting, the medium-term data is missing and therefore, the model is tasked to predict next day's traffic with only one hour's worth of data from the present day. TrafFormer's performance is worsened when only past-hour data is used, as there is no information about the daily traffic fluctuations. 

\noindent\textbf{Day only}. Here we only use the past-day data $X'$ as input. Compared to the hour-only experiment, the day-only results are worse than hour-only, especially in the METR-LA experiments. In our experiments, the past data is provided as a sequence of traffic speed readings with a two-hour gap between them. This gap causes large fluctuations in the input training sequences, which contributes to the model's inability to learn and therefore perform well. Day only information is not suitable to be used as the main input and better serve as supplementary information to accompany the hour only information. 

\noindent\textbf{Cyclical features}. For this experiment, we omit the time-day embedding matrices. Instead, we transform the day and time features from $X$ and $X'$ to cyclical features. These features are static and non-trainable unlike the embedding vectors. For a feature $x$, we transform it to two values: 

\begin{equation} 
    \begin{aligned}
        x_{sin} = \left(sin\left(\frac{x \times 2 \pi}{max(x)}\right) + 1\right) \times \frac{1}{2} \\
        x_{cos} = \left(cos\left(\frac{x \times 2 \pi}{max(x)}\right) + 1\right) \times \frac{1}{2}
    \end{aligned}
\end{equation}

Where $max(x)$ denotes the maximum possible value for that feature, which is 288 for time data and 7 for the day data. As the name suggests, cyclical features are suitable for features that work in cycles such as time. If we model this feature linearly, for instance by scaling all possible values into the range between 0 and 1, timestamps that should be close to each other--such as 23:58 and 00:00--will be given a very large distance despite their closeness. Cyclical features will fix this issue. 

The results show poor performance on both datasets. The purpose of using the time-day embedding is to enable TrafFormer to learn the correlation between timestamps, both close and far, in the context of the task at hand. Cyclical features will assign similar values to timestamp or day that are close to each other, but for long-term prediction, we are not only interested in timestamps or days close to each other, but rather the full, complex correlation between all input timestamps. Traffic flow from a time stamp may be strongly correlated to timestamps farther away and this effect cannot be captured by cyclical features as they are static. 

\noindent\textbf{Speed only}. Here we discard all time and day features and use only the traffic speed reading from $X$ and $X'$. Experiment results show that using speed only perform poorly, on par with using cyclical features. These two experiments shows the importance of using the embeddings to learn the temporal correlation. 
\section{Conclusion}\label{sec:6_conclusion}

In this paper, we explore the novel problem of long-term traffic prediction. From our literature review, we observe that the existing literature predict short-term traffic only, with most being limited to hour-ahead prediction. We address this research gap by exploring the topic of day-ahead traffic prediction. For this task, the commonly used sequence autoencoder model suffers from several deficiencies that limit its effectiveness in long-term prediction. Therefore, we propose the usage of Transformers. We also imbue our model with the novel spatio-temporal multi-head attention module and the time-day embedding modules. Experiments show that our model overall performs the best compared to these baselines. The outcome of this research shows the potential of traffic prediction in assisting traffic planners with proactive measures.

\bibliographystyle{unsrtnat}
\bibliography{references} 

\begin{thebibliography}{42}
\providecommand{\natexlab}[1]{#1}
\providecommand{\url}[1]{\texttt{#1}}
\expandafter\ifx\csname urlstyle\endcsname\relax
  \providecommand{\doi}[1]{doi: #1}\else
  \providecommand{\doi}{doi: \begingroup \urlstyle{rm}\Url}\fi

\bibitem[Cosgove(2015)]{cosgove2015traffic}
David Cosgove.
\newblock \emph{Traffic and congestion cost trends for Australian capital
  cities}.
\newblock Department of Infrastructure and Regional Development, Bureau of
  Infrastructure, Transport and Regional Economics, Canberra, 2015.

\bibitem[BBC(2017)]{BBC2017}
Singapore to freeze car numbers.
\newblock \url{https://www.bbc.com/news/business-41730778}, 2017.
\newblock Accessed: 20 November 2018.

\bibitem[Carnegie and Voorhees(2016)]{Carnegie2016}
Jon Carnegie and Alan~M. Voorhees.
\newblock The cost of roadway construction, operations and maintenance in new
  jersey.
\newblock Technical report, Rugers University, 2016.

\bibitem[Yu et~al.(2017)Yu, Wu, Wang, Wang, and Ma]{YuHaiyang2017}
Haiyang Yu, Zhihai Wu, Shuqin Wang, Yunpeng Wang, and Xiaolei Ma.
\newblock {Spatiotemporal recurrent convolutional networks for traffic
  prediction in transportation networks}.
\newblock \emph{Sensors}, 17\penalty0 (7):\penalty0 1--16, 2017.

\bibitem[Huang et~al.(2014)Huang, Song, Hong, and Xie]{Huang2014}
Wenhao Huang, Guojie Song, Haikun Hong, and Kunqing Xie.
\newblock Deep architecture for traffic flow prediction: Deep belief networks
  with multitask learning.
\newblock \emph{IEEE Transactions on Intelligent Transportation Systems},
  15:\penalty0 2191--2201, 10 2014.

\bibitem[Wu and Tan(2016)]{Wu2016}
Yuankai Wu and Huachun Tan.
\newblock Short-term traffic flow forecasting with spatial-temporal correlation
  in a hybrid deep learning framework.
\newblock \emph{CoRR}, 1612.01022, 2016.

\bibitem[Cheng et~al.(2017)Cheng, Zhang, Zhou, and Xu]{Cheng2017}
Xingyi Cheng, Ruiqing Zhang, Jie Zhou, and Wei Xu.
\newblock Deeptransport: Learning spatial-temporal dependency for traffic
  condition forecasting.
\newblock \emph{CoRR}, 1709.09585, 2017.

\bibitem[Dai et~al.(2017)Dai, Fu, Lin, Li, and Wang]{Dai2017}
Xingyuan Dai, Rui Fu, Yilun Lin, Li~Li, and Fei{-}Yue Wang.
\newblock Deeptrend: {A} deep hierarchical neural network for traffic flow
  prediction.
\newblock \emph{CoRR}, 1707.03213, 2017.

\bibitem[S.~Ahmed and R.~Cook(1979)]{Ahmed1979}
Mohamed S.~Ahmed and Allen R.~Cook.
\newblock Analysis of freeway traffic time series data by using box-jenkins
  techniques.
\newblock \emph{Transportation Research Record}, 773:\penalty0 1--9, 01 1979.

\bibitem[Lee and Fambro(1999)]{Lee1999}
Sangsoo Lee and Daniel Fambro.
\newblock Application of subset autoregressive integrated moving average model
  for short-term freeway traffic volume forecasting.
\newblock \emph{Transportation Research Record}, 1678:\penalty0 179--188, 11
  1999.

\bibitem[Williams(2001)]{Williams2001}
Billy Williams.
\newblock Multivariate vehicular traffic flow prediction: evaluation of arimax
  modeling.
\newblock \emph{Journal of the Transportation Research Board}, 1776:\penalty0
  194--200, 2001.

\bibitem[Williams and A.~Hoel(2003)]{Williams2003}
Billy Williams and Lester A.~Hoel.
\newblock Modeling and forecasting vehicular traffic flow as a seasonal arima
  process: Theoretical basis and empirical results.
\newblock \emph{Journal of Transportation Engineering}, 129:\penalty0 664--672,
  11 2003.

\bibitem[Kamarianakis and Prastacos(2003)]{Kamarianakis2002}
Yiannis Kamarianakis and Poulicos Prastacos.
\newblock Forecasting traffic flow conditions in an urban network : Comparison
  of multivariate and univariate approaches.
\newblock \emph{Journal of the Transportation Research Board}, pages 74--84,
  2003.

\bibitem[Karlaftis and Vlahogianni(2011)]{Karlaftis2011}
M.G. Karlaftis and E.I. Vlahogianni.
\newblock Statistical methods versus neural networks in transportation
  research: Differences, similarities and some insights.
\newblock \emph{Transportation Research Part C: Emerging Technologies},
  19\penalty0 (3):\penalty0 387 -- 399, 2011.

\bibitem[Li et~al.(2018)Li, Zhao, Cong, Jensen, and Wei]{Li2018}
Xiucheng Li, Kaiqi Zhao, Gao Cong, Christian~S. Jensen, and Wei Wei.
\newblock Deep representation learning for trajectory similarity computation.
\newblock In \emph{2018 IEEE 34th International International Conference on
  Data Engineering}, pages 617--628, 2018.

\bibitem[Dougherty et~al.(1993)Dougherty, Kirby, and Boyle]{Dougherty1993}
M.S. Dougherty, H.R. Kirby, and R.D. Boyle.
\newblock The use of neural networks to recognise and predict traffic
  congestion.
\newblock \emph{Traffic Engineering and Control}, 34\penalty0 (6):\penalty0
  311--314, 01 1993.

\bibitem[Vlahogianni et~al.(2005)Vlahogianni, Karlaftis, and
  Golias]{Vlahogianni2005}
Eleni~I. Vlahogianni, Matthew~G. Karlaftis, and John~C. Golias.
\newblock Optimized and meta-optimized neural networks for short-term traffic
  flow prediction: A genetic approach.
\newblock \emph{Transportation Research Part C: Emerging Technologies},
  13\penalty0 (3):\penalty0 211 -- 234, 2005.

\bibitem[Zheng and Lee(2006)]{Zheng2006}
Weizhong Zheng and Der-Horng Lee.
\newblock Short-term freeway traffic flow prediction: Bayesian combined neural
  network approach.
\newblock \emph{Journal of Transportation Engineering}, 132, 2 2006.

\bibitem[Zhong et~al.(2005)Zhong, Sharma, and Lingras]{Zhong2005}
Ming Zhong, Satish Sharma, and Pawan Lingras.
\newblock Short-term traffic prediction on different types of roads with
  genetically designed regression and time delay neural network models.
\newblock \emph{Journal of Computing in Civil Engineering}, 19\penalty0
  (1):\penalty0 94--103, 2005.

\bibitem[Chan et~al.(2012)Chan, S.~Dillon, Singh, and Chang]{Chan2012}
Kit~Yan Chan, Tharam S.~Dillon, Jaipal Singh, and Elizabeth Chang.
\newblock Neural network based models for short-term traffic flow forecasting
  using a hybrid exponential smoothing and {L}evenberg {M}arquardt algorithm.
\newblock \emph{IEEE Transactions on Intelligent Transportation Systems},
  13:\penalty0 644--654, 06 2012.

\bibitem[Yisheng et~al.(2015)Yisheng, Duan, and Kang]{Yisheng2015}
Lv~Yisheng, Yanjie Duan, and Wenwen Kang.
\newblock {Traffic Flow Prediction With Big Data : A Deep Learning Approach}.
\newblock \emph{IEEE Transactions on Intelligent Transportation Systems},
  16\penalty0 (2):\penalty0 865--873, 2015.

\bibitem[Ma et~al.(2015)Ma, Tao, Wang, Yu, and Wang]{Ma2015}
Xiaolei Ma, Zhimin Tao, Yinhai Wang, Haiyang Yu, and Yunpeng Wang.
\newblock Long short-term memory neural network for traffic speed prediction
  using remote microwave sensor data.
\newblock \emph{Transportation Research Part C: Emerging Technologies},
  54:\penalty0 187 -- 197, 2015.

\bibitem[Tian and Pan(2015)]{Tian2015}
Y.~Tian and L.~Pan.
\newblock Predicting short-term traffic flow by long short-term memory
  recurrent neural network.
\newblock \emph{IEEE International Conference on Smart
  City/SocialCom/SustainCom (SmartCity)}, pages 153--158, 2015.

\bibitem[Fouladgar et~al.(2017)Fouladgar, Parchami, Elmasri, and
  Ghaderi]{Fouladgar2017}
Mohammadhani Fouladgar, Mostafa Parchami, Ramez Elmasri, and Amir Ghaderi.
\newblock {Scalable Deep Traffic Flow Neural Networks for Urban Traffic
  Congestion Prediction}.
\newblock \emph{International Joint Conference on Neural Networks}, pages
  2251--2258, 2017.

\bibitem[Kang et~al.(2017)Kang, Yisheng, and Chen]{Kang2017}
Danqing Kang, Lv~Yisheng, and Yuan-yuan Chen.
\newblock {Short-term traffic flow prediction with LSTM recurrent neural
  network}.
\newblock \emph{Intelligent Transportation Systems}, pages 1--6, 10 2017.

\bibitem[Cho et~al.(2014)Cho, van Merrienboer, G{\"{u}}l{\c{c}}ehre, Bougares,
  Schwenk, and Bengio]{Cho2014}
Kyunghyun Cho, Bart van Merrienboer, {\c{C}}aglar G{\"{u}}l{\c{c}}ehre, Fethi
  Bougares, Holger Schwenk, and Yoshua Bengio.
\newblock Learning phrase representations using {RNN} encoder-decoder for
  statistical machine translation.
\newblock \emph{CoRR}, abs/1406.1078, 2014.

\bibitem[Sutskever et~al.(2014)Sutskever, Vinyals, and
  Le]{sutskever2014sequence}
Ilya Sutskever, Oriol Vinyals, and Quoc~V Le.
\newblock Sequence to sequence learning with neural networks.
\newblock \emph{Advances in neural information processing systems}, 27, 2014.

\bibitem[He et~al.(2019)He, Chow, and Zhang]{He2019}
Zhixiang He, Chi-Yin Chow, and Jia-Dong Zhang.
\newblock Stann: A spatio-temporal attentive neural network for traffic
  prediction.
\newblock \emph{IEEE Access}, pages 4795--4806, 2019.

\bibitem[Do et~al.(2019)Do, Vu, Vo, Liu, and Phung]{Do2019}
Loan Do, Hai Vu, Bao Vo, Zhiyuan Liu, and Dinh Phung.
\newblock An effective spatial-temporal attention based neural network for
  traffic flow prediction.
\newblock \emph{Transportation Research Part C: Emerging Technologies},
  108:\penalty0 12--28, 2019.

\bibitem[Pan et~al.(2019)Pan, Liang, Wang, Yu, Zheng, and Zhang]{Pan2019}
Zheyi Pan, Yuxuan Liang, Weifeng Wang, Yong Yu, Yu~Zheng, and Junbo Zhang.
\newblock Urban traffic prediction from spatio-temporal data using deep meta
  learning.
\newblock \emph{Proceedings of the 25th ACM SIGKDD International Conference on
  Knowledge Discovery and Data Mining}, pages 1720–--1730, 2019.

\bibitem[Li et~al.(2022)Li, Feng, Yan, Jin, Yang, Sun, Jin, and
  Li]{Li2022Dynamic}
Fuxian Li, Jie Feng, Huan Yan, Guangyin Jin, Fan Yang, Funing Sun, Depeng Jin,
  and Yong Li.
\newblock Dynamic graph convolutional recurrent network for traffic prediction:
  Benchmark and solution.
\newblock \emph{ACM Transactions on Knowledge Discovery from Data}, 2022.

\bibitem[Tedjopurnomo et~al.(2022)Tedjopurnomo, Bao, Zheng, Choudhury, and
  Qin]{Tedjopurnomo2022Survey}
David~Alexander Tedjopurnomo, Zhifeng Bao, Baihua Zheng, Farhana~Murtaza
  Choudhury, and A.~K. Qin.
\newblock A survey on modern deep neural network for traffic prediction:
  Trends, methods and challenges.
\newblock \emph{IEEE Transactions on Knowledge and Data Engineering},
  34\penalty0 (4):\penalty0 1544--1561, 2022.

\bibitem[Vaswani et~al.(2017)Vaswani, Shazeer, Parmar, Uszkoreit, Jones, Gomez,
  Kaiser, and Polosukhin]{vaswani2017attention}
Ashish Vaswani, Noam Shazeer, Niki Parmar, Jakob Uszkoreit, Llion Jones,
  Aidan~N Gomez, {\L}ukasz Kaiser, and Illia Polosukhin.
\newblock Attention is all you need.
\newblock \emph{Advances in neural information processing systems}, 30, 2017.

\bibitem[Jagadish et~al.(2014)Jagadish, Gehrke, Labrinidis, Papakonstantinou,
  Patel, Ramakrishnan, and Shahabi]{Jagadish2014Big}
H.~V. Jagadish, Johannes Gehrke, Alexandros Labrinidis, Yannis
  Papakonstantinou, Jignesh~M. Patel, Raghu Ramakrishnan, and Cyrus Shahabi.
\newblock Big data and its technical challenges.
\newblock \emph{Communications of the ACM}, 57\penalty0 (7):\penalty0 86–94,
  2014.

\bibitem[Li et~al.(2017)Li, Yu, Shahabi, and Liu]{Li2017}
Yaguang Li, Rose Yu, Cyrus Shahabi, and Yan Liu.
\newblock Diffusion convolutional recurrent neural network: Data-driven traffic
  forecasting.
\newblock \emph{CoRR}, 1707.01926, 2017.

\bibitem[Chen et~al.(2019)Chen, Li, Teo, Zou, Wang, Wang, and
  Zeng]{Chen2019Gated}
Cen Chen, Kenli Li, Sin~G Teo, Xiaofeng Zou, Kang Wang, Jie Wang, and Zeng
  Zeng.
\newblock Gated residual recurrent graph neural networks for traffic
  prediction.
\newblock \emph{Proceedings of the AAAI conference on artificial intelligence},
  33\penalty0 (01):\penalty0 485--492, 2019.

\bibitem[Chen et~al.(2020)Chen, Chen, Lai, Jin, Liu, Li, Wei, Wang, Tang,
  Huang, and Hua]{Chen2020Dynamic}
Ken Chen, Fei Chen, Baisheng Lai, Zhongming Jin, Yong Liu, Kai Li, Long Wei,
  Pengfei Wang, Yandong Tang, Jianqiang Huang, and Xian-Sheng Hua.
\newblock Dynamic spatio-temporal graph-based cnns for traffic flow prediction.
\newblock \emph{IEEE Access}, 8:\penalty0 185136--185145, 01 2020.

\bibitem[Wang et~al.(2020)Wang, Ma, Wang, Jin, Wang, Tang, Jia, and
  Yu]{Wang2020Traffic}
Xiaoyang Wang, Yao Ma, Yiqi Wang, Wei Jin, Xin Wang, Jiliang Tang, Caiyan Jia,
  and Jian Yu.
\newblock Traffic flow prediction via spatial temporal graph neural network.
\newblock In \emph{Proceedings of The Web Conference 2020}, page 1082–1092,
  2020.

\bibitem[Huang et~al.(2021)Huang, Kong, Dai, and Liu]{Huang2021LTPHM}
Chuyin Huang, Weiyang Kong, Genan Dai, and Yubao Liu.
\newblock Ltphm: Long-term traffic prediction based on hybrid model.
\newblock In \emph{Proceedings of the 30th ACM International Conference on
  Information and Knowledge Management}, page 3093–3097, 2021.

\bibitem[Smith(2015)]{Smith2015Cyclical}
Leslie~N. Smith.
\newblock Cyclical learning rates for training neural networks, 2015.

\bibitem[Yu et~al.(2018)Yu, Yin, and Zhu]{Yu2018}
Bing Yu, Haoteng Yin, and Zhanxing Zhu.
\newblock Spatio-temporal graph convolutional neural network: {A} deep learning
  framework for traffic forecasting.
\newblock \emph{International Joint Conference on Artificial Intelligence},
  pages 3634--3640, 2018.

\bibitem[Qi et~al.(2022)Qi, Zhao, Tanin, Cui, Nassir, and Sarvi]{qi2022gamcn}
Jianzhong Qi, Zhuowei Zhao, Egemen Tanin, Tingru Cui, Neema Nassir, and Majid
  Sarvi.
\newblock A graph and attentive multi-path convolutional network for traffic
  prediction.
\newblock In \emph{IEEE Transactions on Knowledge and Data Engineering}, 2022.

\end{thebibliography}

\end{document}